\documentclass{article}

\PassOptionsToPackage{numbers, compress}{natbib}

\usepackage[final]{neurips_2021}

\usepackage{microtype}
\usepackage{graphicx}
\usepackage{subcaption}
\usepackage{booktabs} 
\usepackage{cancel}
\usepackage{wrapfig}
\usepackage[table,dvipsnames]{xcolor}
\usepackage[titlenumbered, ruled]{algorithm2e}
\usepackage{xr-hyper}
\usepackage{hyperref}

\usepackage{nicefrac}       
\usepackage{amssymb,amsfonts}       
\usepackage{amsmath,stackengine,mathtools}
\usepackage{amsthm}

\DeclareMathOperator*{\argmax}{arg\,max} 
\DeclareMathOperator*{\argmin}{arg\,min}

\theoremstyle{definition}
\newtheorem{definition}{Definition}

\theoremstyle{definition}

\theoremstyle{definition}
\newtheorem{remark}{Remark}[section]

\theoremstyle{definition}
\newtheorem{assumptions}{Assumptions}

\usepackage[utf8]{inputenc}

\usepackage{xspace}
\usepackage[T1]{fontenc}    

\usepackage{cleveref}
\crefrangelabelformat{equation}{(#3#1#4--#5#2#6)}
\crefname{equation}{eq.}{eqs.}
\Crefname{equation}{Equation}{Equations}
\usepackage{diagbox}
\crefformat{section}{\S#2#1#3}
\crefformat{subsection}{\S#2#1#3}
\crefformat{subsubsection}{\S#2#1#3}

\usepackage{todonotes}

\usepackage{enumitem}
\setlist[itemize]{leftmargin=*}
\usepackage{multirow}

\usepackage{dsfont}
\newcommand{\mat}[1]{\mathbf{#1}}

\newcommand{\dataset}{\mathcal{D}}

\newcommand{\expectation}[2]{ \mathbb{E}_{#1}{\left[#2\right]}}

\newcommand{\DInt}{\dataset^I}
\newcommand{\DObs}{\dataset^O}

\newcommand{\graph}{\mathcal{G}}
\newcommand{\scmdef}{\left\langle \mat{U},\mat{V}, P(\mat{U}),\mat{F} \right\rangle}

\newcommand{\DO}[2]{\operatorname{do} \!  \left(#1 = #2\right)}

\newcommand{\win}{t_{i-1}:t_i}
\newcommand{\scmt}{\mathcal{M}_i}
\newcommand{\grapht}{\mathcal{G}_i(\scmt)}
\newcommand{\ind}{\mathds{1}_{i>0}}
\newcommand{\IPrev}{I_{i-1}}
\newcommand{\meanreward}{\expectation{\pi}{Y_n(A_n) \mid \ind \cdot I_{i-1}}}
\newcommand{\pom}{\mathbb{A}}
\newcommand{\system}{\mathcal{F}}
\newcommand{\Fhat}{\widehat{\mat{F}}_{\tau}}

\newcommand{\acro}[1]{\textsc{#1}\xspace}

\newcommand{\scmmab}{\acro{scm-mab}}

\newcommand{\mab}{\acro{mab}}

\newcommand{\DAG}{\acro{dag}}

\newcommand{\pomis}{\acro{pomis}}

\newcommand{\sem}{\acro{scm}}

\newcommand{\dbn}{\acro{dbn}}

\newcommand{\ccb}{\acro{ccb}}
\newcommand{\pmf}{\acro{pmf}}
\newcommand{\mdp}{\acro{mdp}}

\title{Chronological Causal Bandits}

\author{
Neil Dhir\\
The Alan Turing Institute\\
\texttt{ndhir@turing.ac.uk} \\
}

\begin{document}

\maketitle

\begin{abstract}
This paper studies an instance of the multi-armed bandit (\mab) problem, specifically where several causal \textsc{mab}s operate chronologically in the same dynamical system. Practically the reward distribution of each bandit is governed by the same non-trivial dependence structure, which is a dynamic causal model. Dynamic because we allow for each causal \mab to depend on the preceding \mab and in doing so are able to transfer information between agents. Our contribution, the Chronological Causal Bandit (\ccb), is useful in discrete decision-making settings where the causal effects are changing across time and can be informed by earlier interventions in the same system. In this paper we present some early findings of the \ccb as demonstrated on a toy problem.
\end{abstract}

\section{Introduction}
\label{sec:introduction}

A dynamical system evolves in time. Examples include weather, financial markets as well as unmanned aerial vehicles \citep{gauthier2021next}. One practical goal is to take decisions within those systems such as deciding which financial instrument to buy, one day after another. The bandit (\mab) paradigm \citep{robbins1952some,chernoff1959sequential} can help with that. In that environment an agent and an environment interact sequentially \citep{lattimore2020bandit}; the agent picks an action and receives a reward from the environment. The agent continues like so, usually for a finite number of plays, with the goal of maximising the total reward. The challenge in this problem stems from the trade-off between exploiting actions with known rewards, versus exploring actions with unknown rewards.

\begin{wrapfigure}{r}{0.6\textwidth}
    \vspace{-1em}
    \centering
    \includegraphics[width=0.6\textwidth]{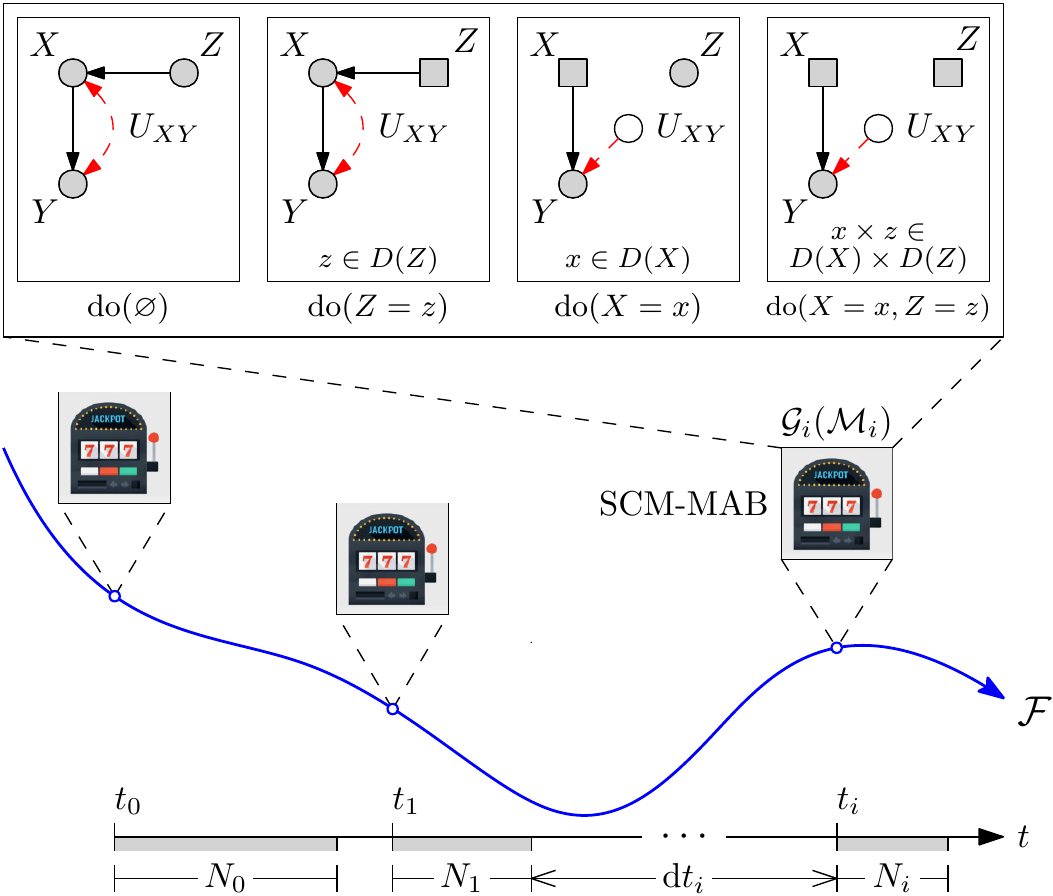}
    \caption{
     Structural causal model framed as a multi-armed bandit problem. \scmmab topology is adapted from figure 3(c) in \citep{lee2018structural}. \textbf{Top}: Each intervention can be construed as pulling an arm and receiving a reward (measuring the causal effect). Shown are all possible interventions -- including sub-optimal ones (e.g. pulling $Z$ and $X$ together). \textbf{Bottom}: The optimal intervention (and consequent samples from the SEM) from the preceding \scmmab are transferred to the current \scmmab, via $\mathcal{F}$. Exogenous variables and incoming edges from $t_{i-1}$ are not shown to avoid clutter. Implemented interventions are represented by squares.}
    \label{fig:sequential_mabs}
    \vspace{-1em}
\end{wrapfigure}

Recently, many studies have addressed the case wherein which there is a non-trivial dependency structure between the arms. One such direction presumes that the dependency structure is modelled explicitly through causal graphs \citep{lee2018structural, lee2019structural, lattimore2016causal, lu2020regret}. We extend that paradigm to also account for the causal temporal dynamics of the system. \textsc{mab}s already constitute sequential decision-making paradigms, but here we expand that idea to cover \emph{chronological} MABs: where the reward distribution is conditional upon the actions taken by earlier MABs (see \cref{fig:sequential_mabs}). We are not considering a Markov Decision Process (\mdp) -- we have no explicit notion of state and consequently do not maintain a model of the state dynamics. This type of transfer learning in causal \textsc{mab}s was also studied by \citep{zhang2017transfer,azar2013sequential} but there the authors transfer information between unrelated tasks whereas we are interested in transfers for when all agents operate in the same (possibly non-stationary) dynamical system. Ours, instead, is more similar to the \emph{restless bandit} \citep{whittle1988restless,guha2010approximation} problem where rewards vary with time (unlike the standard bandit setting where they are fixed but unknown).


\paragraph{Example.}
Consider a dynamical environment $\mathcal{F}$\footnote{We will somewhat abuse standard notation for dynamical systems theory.}, such as a country subject to the COVID-19 pandemic. The highly transient nature of the virus \citep{mandal2020model} necessitates multiple chronological clinical trials, the \emph{start} of each indexed by $t_i$ (see \cref{fig:sequential_mabs}), to find a treatment or vaccine. Suppose we conduct clinical trial $i$ which has $K$ different treatments $\mathcal{A} \triangleq \{a_1,\ldots,a_K\}$ of unknown efficacy for COVID-19 and we have $N_i$ patients in our study group. Patients $\{0,\ldots,n,\ldots, N_i\}$ arrive sequentially, and we must decide on a treatment $A_n \in \mathcal{A}$ to administer to each new patient.

To make this decision, we could learn from how the previous choices of treatments fared for the previous patients. After a sufficient number of trials, we may have a reasonable idea of which treatment is most effective, and from thereon, we could administer that treatment to all patients. However the exploration phase may take a long time and many patients may receive a sub-optimal treatment during that period. But we know than a earlier, similar trial $i-1$, has just concluded and because we are aware of the causal nature of our treatments and their evolution over time, we can condition our trial $i$ on the lessons learned, and actions taken in trial $i-1$, before we start ours. There are two purposes to this (1) the additional information may aid the discovery of the most effective treatment in our trial and (2) it may also show that the most optimal intervention changes over time owing to the highly non-stationary environment of real systems, where a typical assumption \citep{lattimore2020bandit} is for standard \textsc{MAB}s to have a stationary reward distribution \citep{lattimore2020bandit}. The time between two consecutive trials $i$ and $i-1$ is $\mathrm{d}t_i \triangleq t_i - (t_{i-1}+N_{i-1}), \forall i>0$.

\paragraph{Contributions.} The chronological causal bandit (\ccb) extends the \scmmab by \citet{lee2018structural} by conditioning a \scmmab on prior causal bandits played in the same environment $\system$. The result of this is a piece-wise stationary model which offers a novel approach for causal decision making under uncertainty within dynamical systems. The reward process of the arms is non-stationary on the whole, but stationary on intervals \citep{yu2009piecewise}. Specifically, past optimal interventions are transferred across time allowing the present \mab to weigh the utility of those actions in the present game.

\subsection{Preliminaries}
\label{sec:prelims}

\paragraph{Notation.} Random variables are denoted by upper-case letters and their values by lower-case letters. Sets of variables and their values are noted by bold upper-case and lower-case letters respectively. We make extensive use of the do-calculus (for details see \citep[\S 3.4]{pearl2000causality}). Samples (observational) drawn from a system or process unperturbed over time are contained in $\mathcal{D}^O$. Samples (interventional) drawn from a system or process subject to one or more interventions are denoted by $\mathcal{D}^I$. The domain of a variable is denoted by $D(\cdot)$ where e.g. $x \in D(X)$ and $\mat{x} \in D(\mat{X}) \equiv x_1 \times x_2 \times  \ldots \times x_{|\mat{x}|} \in D(X_1) \times D(X_2) \times \ldots \times D(X_{|\mat{x}|})$.

\paragraph{Structural causal model.} Structural causal models (SCMs) \citep[ch. 7]{pearl2000causality} are used as the semantics framework to represent an underlying environment. For the exact definition as used by Pearl see \citep[def. 7.1.1]{pearl2000causality}. Let $\mathcal{M}$ be a \sem parametrised by the quadruple $\scmdef$. Here, $\mat{U}$ is a set of exogenous variables which follow a joint distribution $P(\mat{U})$ and $\mat{V}$ is a set of endogenous (observed) variables. Within $\mat{V}$ we distinguish between two types of variables: manipulative (treatment) and target (output) variables (always denoted by $Y$ in this paper). Further, endogenous variables are determined by a set of functions $\mat{F} \subset \mathcal{F}$. Let $\mat{F} \triangleq \{f_i\}_{V_i \in \mat{V}}$ \citep[\S 1]{lee2020characterizing} s.t. each $f_i$ is a mapping from (the respective domains of) $U_i \cup P\!A_i$ to $V_i$ -- where $U_i \subseteq \mat{U}$ and $P\!A_i \subseteq \mat{V} \setminus V_i$. Graphically, each SCM is associated with a causal diagram (a directed acyclical graph, \DAG for short) $\graph  = \left\langle \mat{V}, \mat{E} \right\rangle$ where the edges are given by $\mat{E}$. Each vertex in the graph corresponds to a variable and the directed edges point from members of $P\!A_i$ and $U_i$ toward $V_i$ \citep[ch. 7]{pearl2000causality}. A directed edge is s.t. $V_i \leftarrow V_j \in \mat{E}$ if $V_i \in P\!A_j$ (i.e. $V_j$ is a child of $V_i$). A bidirected edge between $V_i$ and $V_j$ occurs if they share an unobserved confounder which is to say $\mat{U}_i \cap \mat{U}_j \neq \varnothing$ \citep{lee2018structural}. Unless otherwise stated, from hereon, when referring to $\mat{V}$ we are implicitly considering $\mat{V} \setminus \{Y\}$ -- i.e. the manipulative variables not including the outcome variable. Finally, the fundamental do-operator $\DO{\mat{V}}{\mat{v}}$ represents the operation of fixing a set of endogenous variable(s) $\mat{V}$ to constant value(s) $\mat{v}$ irrespective of their original mechanisms. Throughout we do not consider graphs with non-manipulative variables. For more a more incisive discussion on the properties of SCMs we refer the reader to \citep{pearl2000causality, bareinboim2016causal}.

\paragraph{Multi-armed bandit.} The MAB setting \citep{robbins1952some} entertains a discrete sequential decision-making scenario in which an agent selects an action or `arm' $a \in \mathcal{A}$ according to a policy $\pi$, and receives a stochastic reward $Y(a)$, emanating from an unknown distribution particular to each arm. The expectation of the reward is given by $\mu_a$. The goal of the agent is to optimise the arm selection sequence and thereby maximise the expected reward $\sum ^N _{n=0} \expectation{\pi}{Y(A_n)}$ after $N$ rounds, where $\expectation{\pi}{\cdot}$ is the expectation under the given policy and $A_n$ is the arm played on the $n^{th}$ round. We will use a similar performance measure, the cumulative regret \citep{lee2018structural} $R_N = N \mu^* - \sum^N_{n=1} \expectation{\pi}{Y(A_n)}$ where the max reward is $\mu^* = \max_{a \in \mathcal{A}} \mu_a$ . Using the regret decomposition lemma \citep[Lemma 4.5]{lattimore2020bandit}, we can write this in the form
\begin{equation}
    \label{eq:standard_regret}
    R_N = \sum^K_{a=1} \Delta_a \expectation{\pi}{\#_a(N)}
\end{equation}
where each arm's gap from the best arm (``suboptimality gap'' \citep[\S 4.5]{lattimore2020bandit}) is $\Delta_a = \mu^* - \mu_a$ and $\#_a(N)$ is the total number of times arm $a$ was played after $N$ rounds.

\begin{wrapfigure}{r}{0.35\textwidth}
\vspace{-0.5cm}
    \centering
    \includegraphics[width=0.35\textwidth]{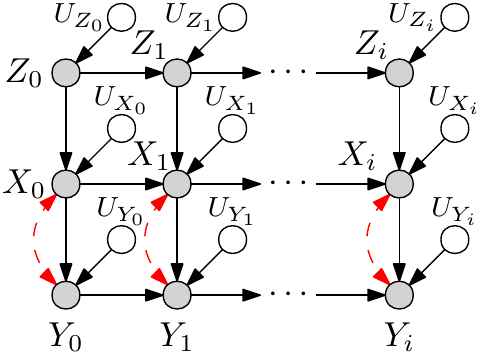}
    \caption{Toy \sem used throughout this paper, based on \citep[Fig. 3(c)]{lee2018structural} but with the important difference that this \sem is propagated in time.}
    \vspace{-0.5cm}
    \label{fig:example_scm}
\end{wrapfigure}

\paragraph{Connecting SCMs to MABs.} Echoing the approach taken by \citet[\S2]{lee2018structural}, using the notation and concepts introduced in \cref{sec:prelims}; let $\mathcal{M}$ be an SCM parametrised by $\left\langle \mat{U},\mat{V}, P(\mat{U}),\mat{F} \right\rangle$ where $Y \in \mat{V}$ is the, as-noted, reward variable. The \emph{arms} of the bandit are given by the set $\mathcal{A} = \{ \mat{a} \in D(\mat{A}) \mid \mat{A} \subseteq \mathcal{P}(\mat{V} \setminus \{Y\}) \}$\footnote{For example if $\mat{A} = \{Z,X\}$ then $D(\mat{A}) = D(Z)\times D(X)$. If $D(Z) = D(X) =[0,1]$ then $D(\mat{A}) = \{(0,0), (0,1),(1,0),(1,1) \}.$}. This is a set of all possible interventions on endogenous variables except the reward variable \citep[\S2]{lee2018structural}. Each arm is associated with a reward distribution $p(Y \mid \mathrm{do}(\mat{a}))$ where the mean reward $\mu_{\mat{a}}$ is $\expectation{\pi}{Y \mid \mathrm{do}(\mat{a})}$. This is the \scmmab setting \citep{lee2018structural}, fully represented by the tuple $\left\langle \mathcal{M}, Y \right\rangle$. As noted by the authors an agent facing a \scmmab $\left\langle \mathcal{M}, Y \right\rangle$, intervenes (plays arms) with knowledge of $\mathcal{G}(\mathcal{M})$ and $Y$ but does not have access to the structural equations model $\mat{F}$ and the joint distribution over the exogenous variables $P(\mat{U})$.

\paragraph{Causality across time.} Taking inspiration from \citet[\S 2]{DCBO}, the authors consider causality in time, manifested by propagating a \DAG in time, and connecting each time-slice \DAG with directed edges as shown in \cref{fig:example_scm}. By doing this we are invariably considering a dynamic Bayesian network (\dbn) \citep{koller2009probabilistic}. As we are making interventions on time-slices of the DBN, we introduce notation to aid the exposition of the method.
\begin{definition}
Let $\scmt$ be the \sem at time step $t_i$ defined as $\scmt = \left\langle \mat{U}_{\tau},\mat{V}_{\tau}, P(\mat{U}_{\tau}),\mat{F}_{\tau} \right\rangle$ for $t>0$. The temporal window covered by the \sem spans $\tau \triangleq \win$ i.e. taking into account only the most recent time-slice, and the actions taken therein. It is also possible to increase the size of $\tau$ to include the entire history i.e. $0:t_i$\footnote{The choice of window size is a difficult one. More information is typically better but we may also subject the model to `stale' information i.e. interventions which are no longer of any relevance or worse, misleading in the present scenario.} as is done in \cref{fig:mab_method}.
\end{definition}

\begin{definition}
Let $\grapht$ \citep[p. 203]{pearl2000causality} be the induced subgraph associated with $\scmt$. In $\grapht$, following the rules of do-calculus \citep[Theorem 3.4.1]{pearl2000causality}, the intervened variable(s) at $t_{i-1}$ have no incoming edges i.e. the $t_{i-1}$ time-slice part of $\grapht$ has been mutilated in accordance with the implemented intervention at $t_{i-1}$.
\end{definition}

\section{Chronological Causal Bandits}

The \scmmab is particular to one graph, wherein which we seek to minimise $R_N$. We instead seek a sequence of interventions which minimise $R_{N_i}$ at each time-step $t_i$ (i.e. the start of the trial)\footnote{For clarity: each trial contains multiple time-trials (rounds), summarised in $N_i$.} of a piece-wise stationary \scmmab as set out in the next paragraph.

\paragraph{Problem statement.}\label{sec:problem_statement}

Similar to \citep{DCBO} the goal of this work is to find a sequence of optimal interventions over time, indexed by $t_i$, by playing a series of sequential conditional \textsc{scm-mab}s or \emph{chronological causal bandits} (\ccb). The agent is provided with $\left\langle \scmt, Y_i \right\rangle$ for each $t_i$ and is then tasked with optimising the arm-selection sequence (within each `trial' $i$) and in so doing, minimise the total regret, for the given horizon $N_{i}$. However, different to the standard \mab problem we must take into account previous interventions as well. We do that by first writing the regret \cref{eq:standard_regret}, in a different form which is conditional on past interventions, enabling the transfer of information
\begin{align}
    \label{eq:intervention_regret}
    R_{N_i} &= \sum^{N_i}_{n=1}  \expectation{\pi}{Y_n(A^*_n) \mid \ind \cdot I_{i-1}} - \meanreward \\
            &\text{ where } A^*_n = \argmax_{a \in \mathcal{A}} \expectation{\pi}{Y_n(a) \mid \ind \cdot I_{i-1}} \nonumber
\end{align}
where $\IPrev = \mathrm{do}(a_{i-1}^*)$ denotes the previously \emph{implemented intervention} and $\ind$ is the indicator function. Now, take particular care w.r.t. the index for which the ``implemented intervention'' concerns. Drawing upon our example in \cref{sec:introduction}, the implemented intervention here corresponds to the treatment that was found to be the most effective during trial $i-1$ (during which $N_{i-1}$ rounds were played). Although the agent finds a sequence of decisions/treatments, only one of them, in this setting, is the overall optimal choice -- i.e. has the lowest, on average, suboptimality gap $\Delta_a$. The implemented intervention at $i-1$ is found by investigating which arm has the lowest regret at the end of horizon $N_{i-1}$
\begin{equation}
    \IPrev = \argmin_{a \in \mathcal{A}}  \{ \Delta_a \expectation{\pi}{\#_a(N_{i-1})} \mid a = 1,\ldots, K \}
\end{equation}
Simply, we pick the arm that has been played the most by the agent. 

\begin{remark}
The above problem statement does not discuss which set of interventions are being considered. Naively one approach is to consider all the $2^{|\mat{V} \setminus \{Y\}|}$ sets in $\mathcal{P}(\mat{V} \setminus \{Y\})$ \citep{cbo}. Though a valid approach, the size will grow exponentially with the complexity of $\grapht$. Intuitively, one may believe that the best and only action worth considering is to intervene on all the parents of the reward variable $Y_i$. This is indeed true provided that $Y_i$ is not confounded by any of its ancestors \citep{lee2018structural}. But whenever unobserved confounders are present (shown by red and dashed edges in \cref{fig:example_scm}) this no longer holds true. By exploiting the rules of do-calculus and the partial-orders amongst subsets \citep{cbo}, \citet{lee2018structural} characterise a reduced set of intervention variables called the possibly optimal minimal intervention set (\pomis) or $\pom$ from hereon\footnote{Generally, $\mathcal{A}$ refers to all $2^{|\mat{V} \setminus \{Y\}|}$ possible interventions whereas $\pom$ refers \emph{only} to the \textsc{pomis}.}, where $\pom \subseteq \mathcal{P}(\mat{V} \setminus \{Y\})$, and typically $|\pom| < |\mathcal{P}(\mat{V} \setminus \{Y\})|$. They demonstrate empirically that the selection of arms based on \textsc{pomis}s make standard \mab solvers ``converge faster to an optimal arm'' \citep[\S 5]{lee2018structural}. Throughout we use the \pomis as well. For the full set of arms for the toy problem in \cref{fig:example_scm}, see \cref{table:arms}.
\end{remark}
\begin{assumptions} \label{assumptions}
To tackle a \ccb style problem, we make the following assumptions (included are also those made by \citet{lee2018structural}  w.r.t. the standard \scmmab):
\begin{enumerate}[leftmargin=*, noitemsep]
    \item Invariance of the causal structure $\mathcal{G}_0(\mathcal{M}_0) = \grapht, \forall i > 0$.
    \item \citet{DCBO} showed that $\mathbb{A}$ does not change across time given assumption (1).
    \item An agent facing a \scmmab $\left\langle \scmt, Y_i \right\rangle$, plays arms with knowledge of $\grapht$ and $Y_i$ but not $P(\mat{U}_{\tau})$ and $\mat{F}_{\tau}$.
\end{enumerate}
\end{assumptions}
Assumption (2) posits that the \DAG is known. If this were not true then we would have to undertake causal discovery (CD) \citep{glymour2019review} or spend the first interactions with the environment \citep{lee2018structural} learning the causal \DAG, from $\DObs$ \citep{spirtes2000causation}, from $\DInt$ \citep{kocaoglu2017experimental} or both \citep{cbo}. As it is, CD is outside the scope of this paper.

\subsection{Transferring information between causal bandits}
\label{sec:transfer}
Herein we describe how to express the reward distribution for trial $i$ as a function of the intervention(s) implemented at the previous trial $i-1$. The key to this section is the relaxation of assumption (3). We seek an estimate $\widehat{\mat{F}}_{\tau}$ for $\mat{F}_{\tau}$ (the true SEM), and $\widehat{P}(\mat{U}_{\tau})$ for $P(\mat{U}_{\tau})$. Thus, if we have $\widehat{\mat{F}}_{\tau}$, we can relay information about past interventions to \scmmab $i$ and thus enable the present reward distribution to take into account those interventions. Reminding ourselves that the members of $\mat{F}_{\tau} \triangleq \{ V_j = f_j(\text{Pa}(j), U_j) \mid j \in \mat{V}_{\tau} \}$. Alas for $i=0$ e.g. we seek function estimates $\{\widehat{f}_{Z_0},\widehat{f}_{X_0},\widehat{f}_{Y_0} \}$. Because $\{ D(V) \subseteq \mathbb{Z} \mid V \in \mat{V}\}$ we model all functions in $\mathcal{M}$ as independent probability mass functions (\pmf).

\begin{figure}[ht!]
    \centering
    \includegraphics[width=\textwidth]{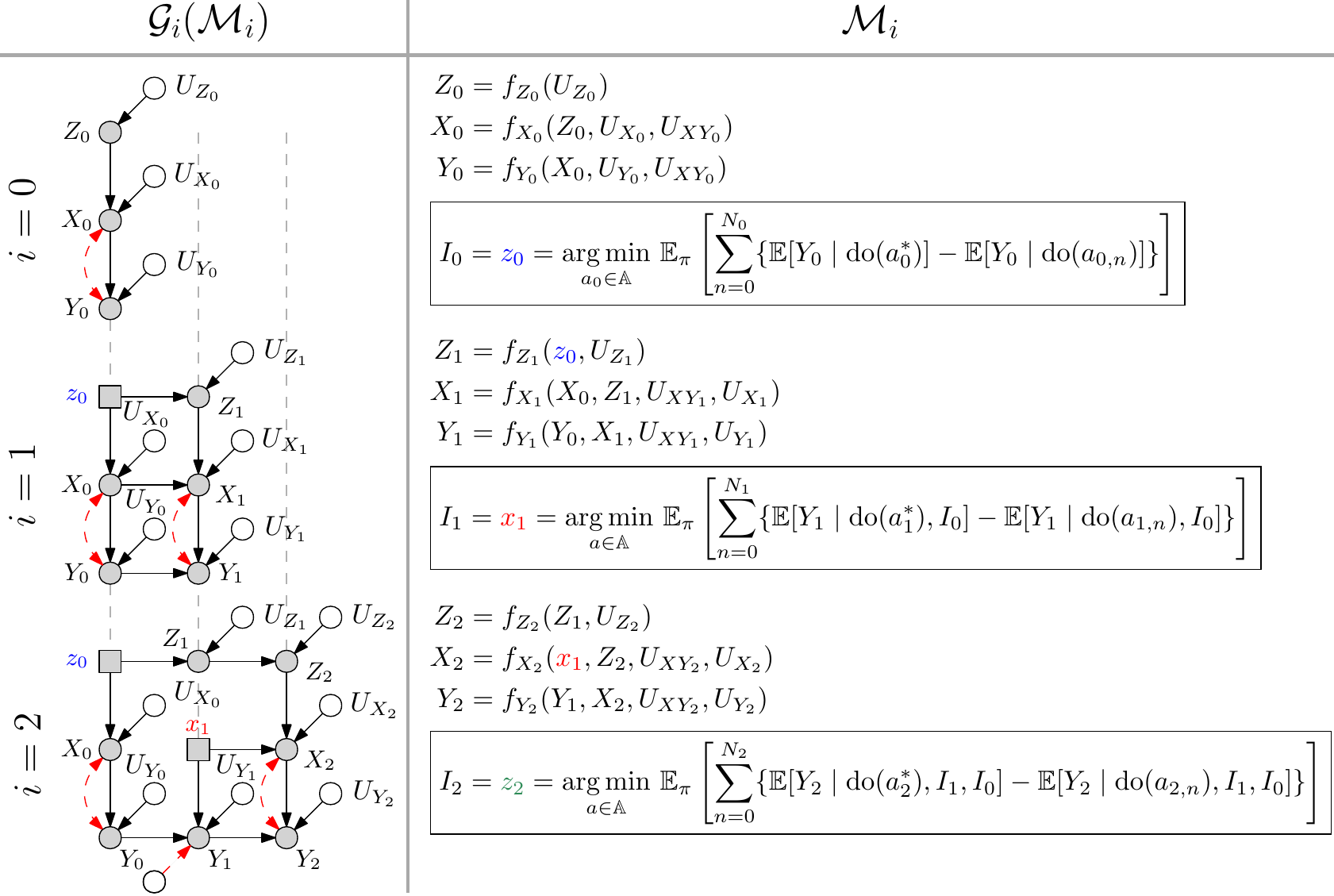}
    \caption{Transfer learning shown through the lens of the \ccb applied to \cref{fig:example_scm}. Each row corresponds to one \scmmab which, if $i>0$, takes into account previous interventions and secondary effects from those interventions (intervening on e.g. $Z_0$ means we can also calculate values for $X_0$ and $Y_0$ in time-slice $t_0$). Further, shown are the structural equation models considered by \ccb at every time step $t \in \{0,1,2\}$. As well as the minimisation task over the cumulative regret, undertaken to find the best action for each time-slice. Square nodes denote intervened variables.
    }
    \label{fig:mab_method}
\end{figure}

\paragraph{Simulation.} Recall that $P(Y \mid \mat{V} = \mat{v})$ is an observational distribution, who's samples are contained in $\DObs$. Where $P(Y \mid \DO{\mat{V}}{\mat{v}} )$ is an interventional distribution, found by fixing variables $\mat{V}$ to $\mat{v}$ in $\mathcal{G}(\mathcal{M})$ -- its samples are contained in $\DInt$. As we are operating in a discrete world, we do not need to approximate any intractable integrals (as is required in e.g. \citep{DCBO, cbo}) to compute the effect of interventions. Rather, by assuming the availability of $\DObs$ and using the do-calculus, we are at liberty to estimate $\mat{F}$ by approximating the individual \pmf{}s that arise from applying the do-calculus (see \citep[Theorem 3.4.1]{pearl2000causality}). Consequently we can build a `simulator' $\Fhat$ using $\DObs$ (which concerns the whole of $\mathcal{M}$, not just the window $\tau$). $\mathcal{D}^{I}$ is very scarce because playing an arm does not yield an observed target value but only a reward (hence we cannot exploit the actions taken during horizon $N_i$). The only interventional data that is available, at each $i$, to each \scmmab $\left\langle \scmt, Y_i \right\rangle$, is the implemented intervention $\IPrev$. The \ccb method is graphically depicted in \cref{fig:mab_method}. 

\section{Experiments}
\label{sec:experiments}

\begin{wrapfigure}{r}{0.35\textwidth}
\vspace{-2em}
    \centering
    \includegraphics[width=0.35\textwidth]{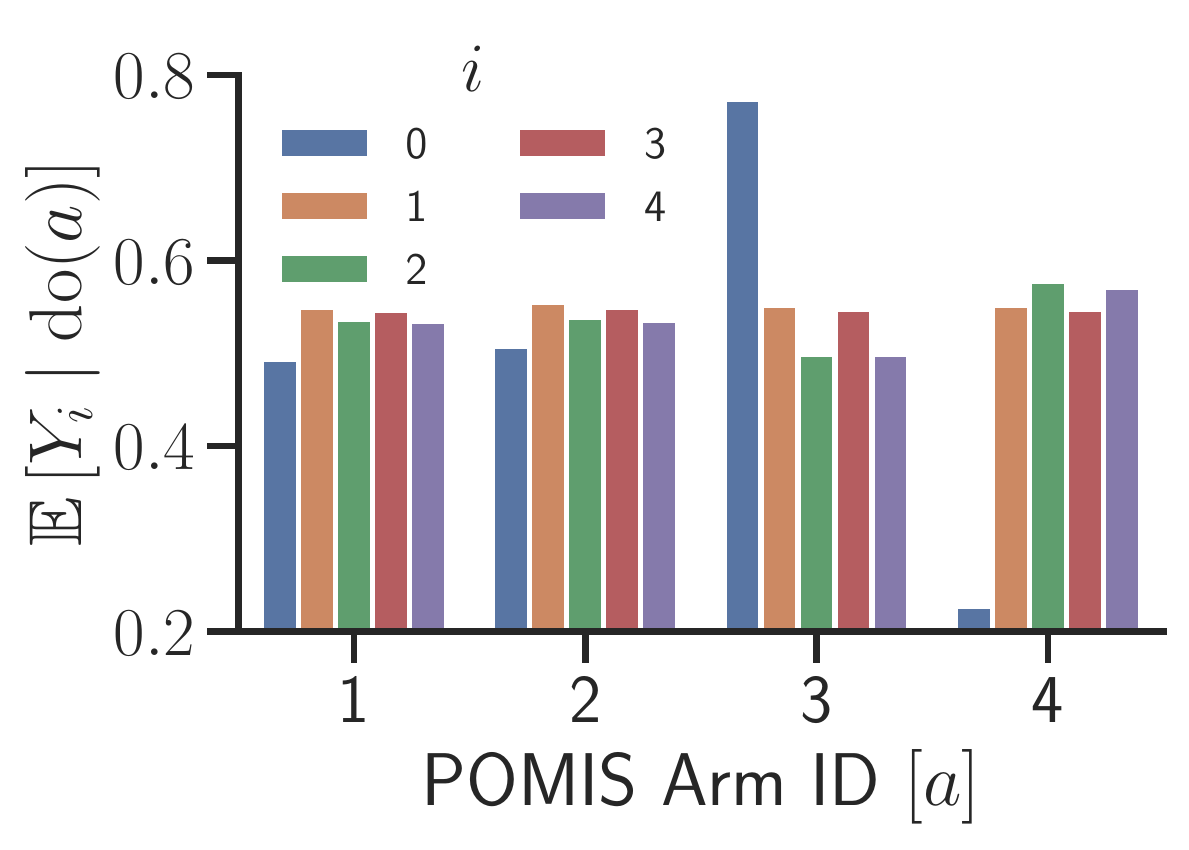}
    \vspace{-2em}
    \caption{Reward distribution for arms in the \pomis across the first five trials $i=0,\ldots,4$. For setting, see \cref{table:arms}.}
    \vspace{-2em}
    \label{fig:reward_distros}
\end{wrapfigure}

We demonstrate some empirical observations on the toy-example used throughout, shown in \cref{fig:example_scm}. We consider the first five time-slices (trials) as shown in \cref{fig:mab_method}. The reward distribution, under the \pomis, for each slice is shown in \cref{fig:reward_distros} (these distributions are found conditional upon the optimal intervention being \emph{implemented} in the preceding trial as shown in \cref{fig:mab_method}). For the true \textsc{sem} see \crefrange{eq:sem_t1a}{eq:sem_t1b} and \crefrange{eq:sem_t0a}{eq:sem_t0b}. \Cref{fig:reward_distros} shows that the system is in effect oscillating in the reward distribution. This is because the optimal intervention changes in-between trials.

\begin{figure}[ht!]
    \centering
    \begin{subfigure}{.33\textwidth}
      \centering
      \includegraphics[width=1.\linewidth]{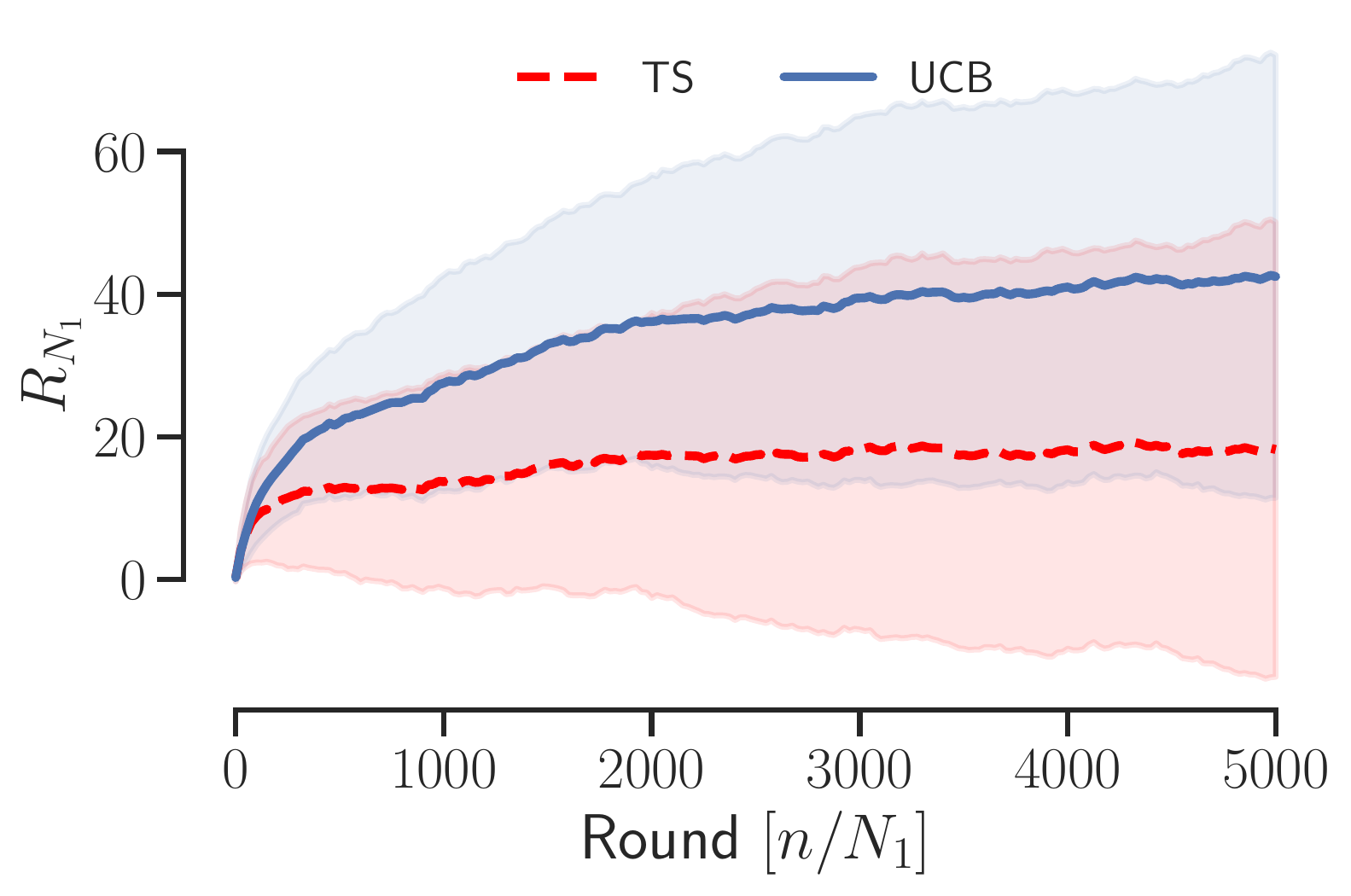}
      \caption{\scmmab at $i=1$. \label{fig:cr_scmmab}}
    \end{subfigure}%
    \begin{subfigure}{.33\textwidth}
      \centering
      \includegraphics[width=1.\linewidth]{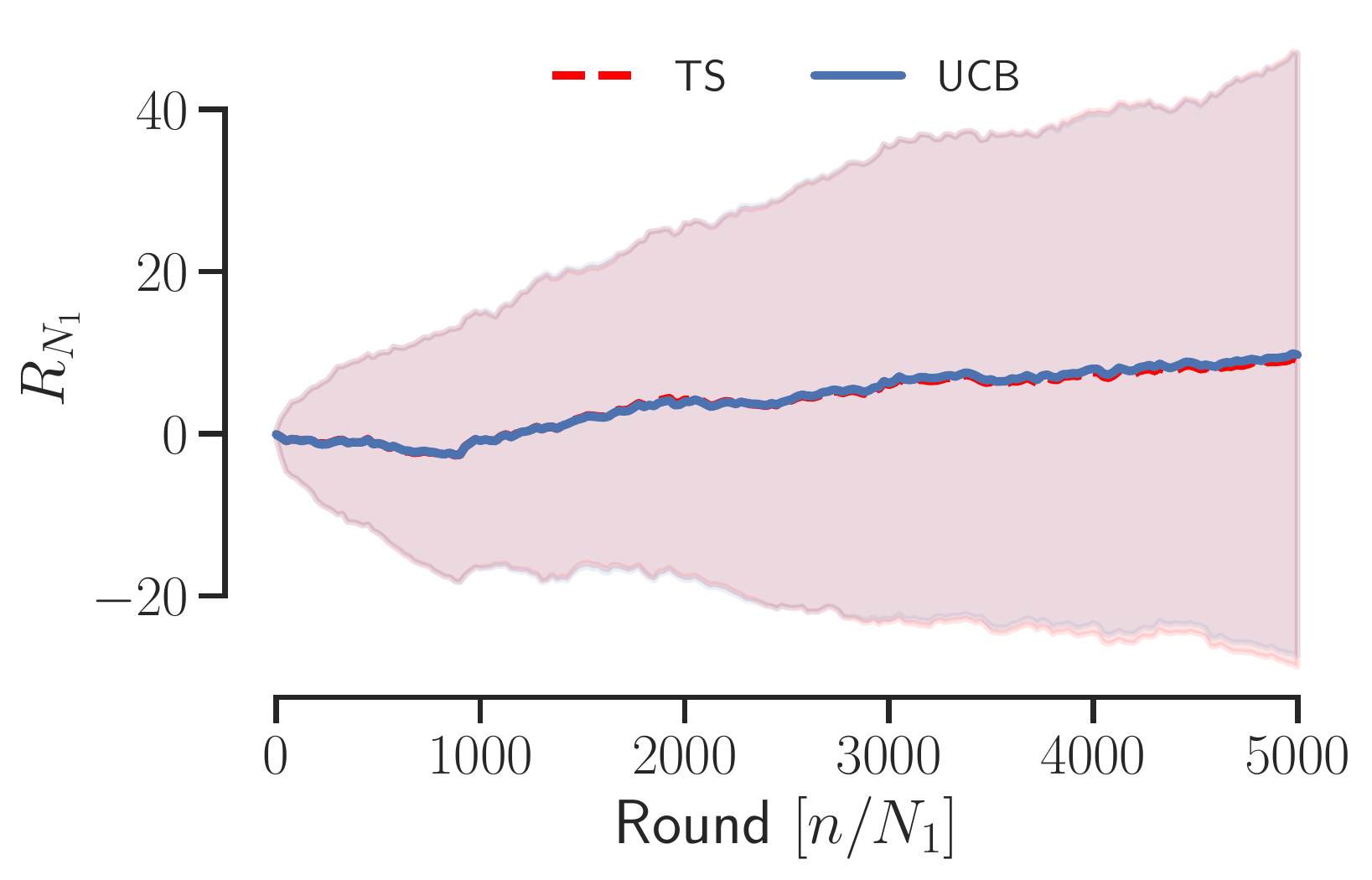}
      \caption{\ccb at $i=1$. \label{fig:cr_ccb}}
    \end{subfigure}%
    \begin{subfigure}{.33\textwidth}
      \centering
      \includegraphics[width=1.\linewidth]{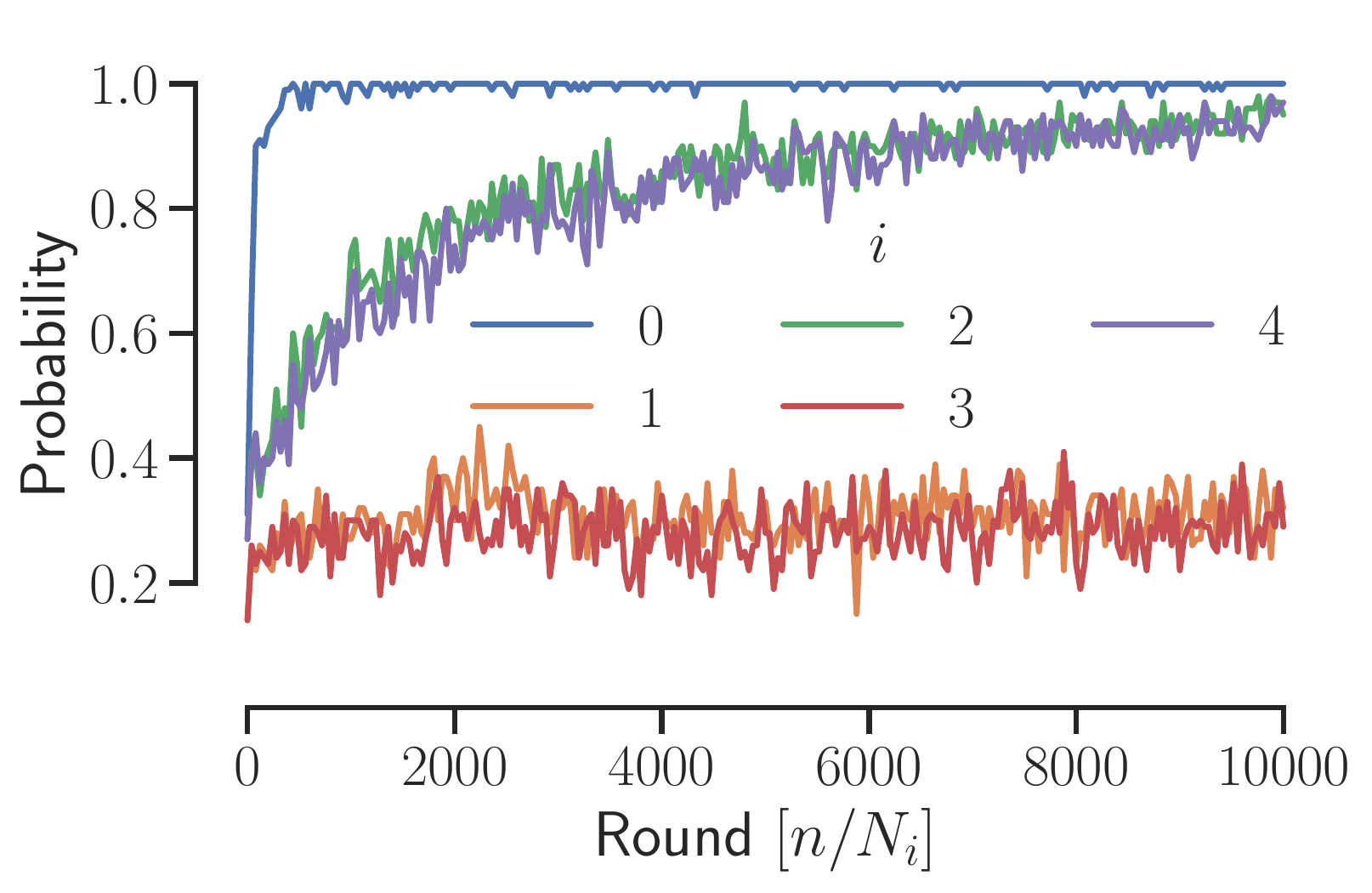}
      \caption{Optimal arm selection proba. \label{fig:TS_results}}
    \end{subfigure}%
    \caption{\Cref{fig:cr_scmmab} and \cref{fig:cr_ccb} (truncated at 5000 rounds) show cumulative regret along with the standard deviation at trial $i=1$. Optimal arm-selection probability is shown in \cref{fig:TS_results}, found using a Thompson sampling policy \citep[\S 5]{lee2018structural}. An equivalent plot for KL-UCB can be found in \cref{fig:UCB_results}.
    \label{fig:results}}
\end{figure}

The horizon for each trial was set to 10,000. We used two common \mab solvers: Thompson sampling (TS, \citep{thompson1933likelihood}) and KL-UCB \citep{cappe2013kullback}. Displayed results are averaged over 100 replicates of each experiment shown in \cref{fig:results}. We investigate the CR and the optimal arm selection probability at various instances in time.

For trial $i=0$, \ccb and \scmmab are targeting the same reward distribution. Consequently they both identify the same optimal intervention $Z_0=z_0$. Come $i=1$ things start to change; having implemented the optimal intervention from $i=0$ \ccb is now targeting a different set of rewards (see \cref{fig:reward_distros}). The \scmmab, being agnostic about past interventions, targets the same reward as previously (blue bars in \cref{fig:reward_distros}). As discussed, this ignores the dynamics of the system; a vaccinated population will greatly alter the healthcare environment, hence to administer the same vaccine (arm 3 at $i=0$) at the next clinical trial ($i=1$), without taking this into account, makes for poor public policy.
\vspace{-1em}

Consider the CR from the \ccb at trial $i=1$ \cref{fig:cr_ccb}; it is lower than the \scmmab in \cref{fig:cr_scmmab}, as it is transferring preceding intervention to the current causal model, \cref{fig:mab_method}($i=1$), and now finds that $X_1=x_1$ is the optimal intervention (see \cref{sec:example_details} for full optimal intervention sequence). Let's now turn to \cref{fig:TS_results}, to minimise the regret the agent should be choosing the optimal action almost all of the time. But it is only possible to reduce regret if the algorithm has discovered the arm with the largest mean. In trials $i=1$ and $i=3$ the reward per arm, across the \pomis, is almost identical. As is stands the agent does not have a reasonable statistical certainty that is has found the optimal arm (orange and red curves in \cref{fig:TS_results}). But all have the same causal effect, why the CR in \cref{fig:cr_ccb} is low.

\section{Conclusion}

We propose the chronological causal bandit (\ccb) algorithm which transfers information between causal bandits which have been played in the dynamical system at an earlier point in time. Some initial findings are demonstrated on a simple toy example where we show that taking the system dynamics into account has a profound effect on the final action selection. Further, whilst we in this example have assumed that $\text{d}t_i$  is the same for each trial, it remains a large assumption that will be studied further in future work. There are many other interesting avenues for further work such as the optimal length of horizon $N_i$ as well as determining the best time $t^*_i$ to play a bandit (i.e. start a trial). Moreover, the current framework allows for confounders to change across time-step -- something we have yet to explore. 

\newpage
\bibliographystyle{icml2021}
\bibliography{references}

\newpage
\appendix
\section{Toy example details}
\label{sec:example_details}

Material herein is adapted from \citep[task 2]{lee2018structural}.

\subsection{Structural equation model}

Generating model for exogenous variables:
\begin{align*}
    \{P(U_{Z_t} = 1) = 0.6 &\mid t = 0,\ldots,T \}\\
    \{P(U_{X_t} = 1) = 0.11 &\mid t = 0,\ldots,T \}\\
    \{P(U_{XY_t} = 1) = 0.51 &\mid t = 0,\ldots,T \}\\
    \{P(U_{Y_t} = 1) = 0.15 &\mid t = 0,\ldots,T \}
\end{align*}
In the \ccb setting, these probabilities remain the same for all time-slices indexed by $\{0,\ldots,t,\ldots,T\}$, as shown by the functions in $\scmt$, when $t>0$:
\begin{align}
    f_Z(u_{Z_t},z_{t-1}) &= u_{Z_t} \wedge z_{t-1} \label{eq:sem_t1a}\\
    f_X(z_t,u_{X_t},u_{XY_t},x_{t-1}) &= u_{X_t} \oplus u_{XY_t} \oplus z_t \oplus x_{t-1}\\
    f_Y(x_t,u_{Y_t},u_{XY_t}, y_{t-1}) &= 1 \oplus u_{Y_t} \oplus u_{XY_t} \oplus x \wedge y_{t-1}. \label{eq:sem_t1b}
\end{align}
When $t=0$ we use the original set of structural equation models \citep[appendix D]{lee2018structural}:
\begin{align}
    f_Z(u_{Z_0}) &= u_{Z_0} \label{eq:sem_t0a}\\
    f_X(z_0,u_{X_0},u_{XY_0}) &= u_{X_0} \oplus u_{XY_0} \oplus z_0 \\
    f_Y(x_0,u_{Y_0},u_{XY_0}) &= 1 \oplus u_{Y_0} \oplus u_{XY_0} \oplus x_0 \label{eq:sem_t0b}
\end{align}
where $\oplus$ is the exclusive disjunction operator and $\wedge$ is the logical conjunction operator (i.e. `and'). The biggest difference between \crefrange{eq:sem_t1a}{eq:sem_t1b} and \crefrange{eq:sem_t0a}{eq:sem_t0b} is that the former has an explicit dependence on the past. Depending on the implemented intervention at $t-1$ one or both of $\{z_{t-1}, x_{t-1}\}$ will be fixed to the value(s) in $I_{t-1}$.
\subsection{Intervention sets}
\label{sec:arms}

``The task of finding the best arm among all possible arms can be reduced to a search within the \textsc{mis}s'' \citep{lee2018structural}. The \pomis is given by $\mathbb{A} = \{\DO{X}{0}, \DO{X}{1}, \DO{Z}{0}, \DO{Z}{1}\}$.

\begin{table}[ht!]
\centering
\caption{Assume binary domains i.e. $D(X) = \{0,1\}$ and $D(Z) = \{0,1\}$. The first arm (ID 0) does nothing i.e. corresponds to the intervention on the empty set $\textrm{do}(\varnothing)$. Arms which belong to the \pomis arms are highlighted. \label{table:arms}}
\begin{tabular}{ccccc}
    \toprule
    \multirow{2}{*}{Arm ID} &
    \multirow{2}{*}{Domain} &
    \multicolumn{2}{r}{Interventions} \\ \cmidrule(lr){3-5}
     &  & $\textrm{do}(X = x)$& $\textrm{do}(Z = z)$ & $\textrm{do}(X = x, Z = z)$ \\
    \midrule
    0 & $\nexists$ & & & \\
    \rowcolor{ForestGreen!40} 1 & $D(X)$ & $x=0$& &\\
    \rowcolor{ForestGreen!40} 2 & $D(X)$ & $x=1$ & & \\
    \rowcolor{ForestGreen!40} 3 & $D(Z)$ & &$z=0$ & \\
    \rowcolor{ForestGreen!40} 4 & $D(Z)$ & &$z=1$ & \\
    5 & $D(X) \times D(Z)$ & & & $x=0,z=0$\\
    6 & $D(X) \times D(Z)$ & & & $x=0,z=1$\\
    7 & $D(X) \times D(Z)$ & & & $x=1,z=0$\\
    8 & $D(X) \times D(Z)$ & & & $x=1,z=1$\\
    \bottomrule
\end{tabular}
\end{table}

\subsection{Additional results}

The optimal intervention sequence is given by $\{Z_0 = z_0, X_1= x_1, Z_1=z_1,X_1=x_1,Z_1=z_1 \}$. Using the example domain, this translates to $\{0,1,0,1,0\}$.

\begin{figure}[ht!]
    \centering
    \includegraphics[width=0.8\linewidth]{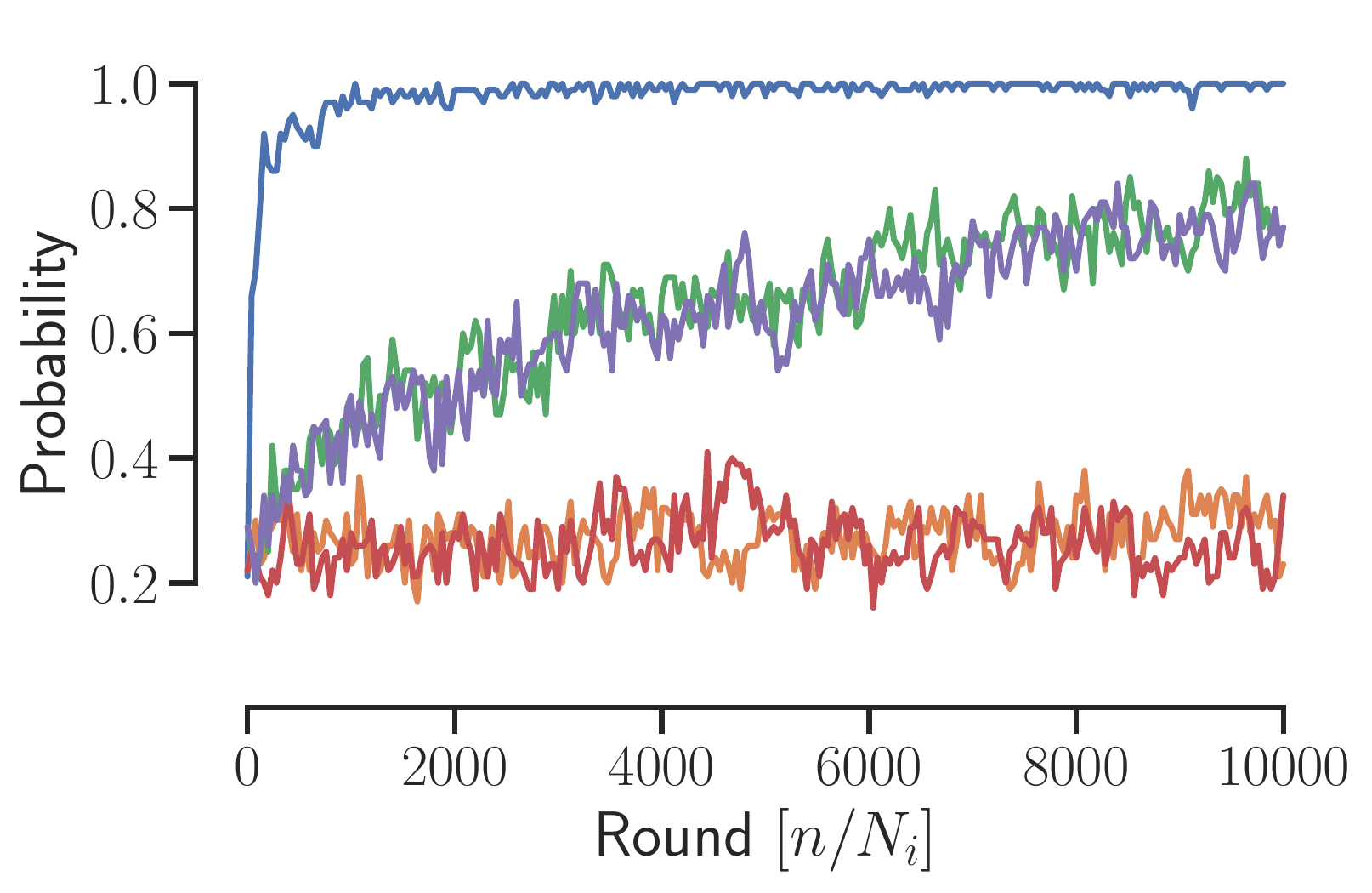}
    \caption{Results using a Kullback-Leibler Upper-confidence bound (KL-UCB) policy $\pi$ \citep{cappe2013kullback}.\label{fig:UCB_results}}
\end{figure}

\end{document}